# A Survey on Techniques of Improving Generalization Ability of Genetic Programming Solutions

Vipul K. Dabhi[1] and Sanjay Chaudhary[2]
[1]Information Technology Department, Dharmsinh Desai University, Nadiad, INDIA.
[2]DA-IICT, Gandhinagar, Gujarat, INDIA

*Abstract*-- **In the field of empirical modeling using Genetic Programming (GP), it is important to evolve solution with good generalization ability. Generalization ability of GP solutions get affected by two important issues: bloat and over-fitting. We surveyed and classified existing literature related to different techniques used by GP research community to deal with these issues. We also point out limitation of these techniques, if any. Moreover, the classification of different bloat control approaches and measures for bloat and over-fitting are also discussed. We believe that this work will be useful to GP practitioners in following ways: (i) to better understand concepts of generalization in GP (ii) comparing existing bloat and over-fitting control techniques and (iii) selecting appropriate approach to improve generalization ability of GP evolved solutions.**

*Index Terms*-- **Genetic Programming, Symbolic Regression, Generalization, Bloat, Over-fitting, Empirical Modeling.**

## I. INTRODUCTION

Different Machine Learning (ML) techniques try to extract implicit relationship that may exist between input variables and output variables of a system. Generally only limited numbers of observations (input-output mappings) is known or available to learner during training phase. Developing a model (solution) from these limited numbers of observations carries a risk of over-fitting. For any ML technique to become trusted, the technique is expected to generate a solution that could achieve same generalization performance on unseen data as obtained on the training data. By generalizing implicit relationship learned during training phase, the success (scalability) of developed solution can be improved. Non-evolutionary ML techniques have realized the importance of generalization and significant research has been done in this area. The issue of generalization ability of evolved Genetic Programming (GP) models has received attention recently and many contributions dealing with the issue have appeared. This paper reviews issues related to and efforts put by researchers to improve generalization performance of GP evolved solutions.

The Minimum Description Length (MDL) [24] approach to improve generalization ability of solutions induced by GP suggests promoting evolution of simpler solutions compare to complex solutions. The approach suggests that it is more likely that complex solutions may contain specific information from training data and thus may overfit it compared to simpler solutions. However, GP practitioners noticed that average size of solutions increases very quickly after a certain number of generations, not matched by any corresponding gain in fitness. This phenomenon of increase in solution size without significant gain in terms of fitness is known as bloat. MDL principle suggests that over-fitting and size of solution are related entities. However, recent contribution [31] show that bloat and over-fitting are two independent phenomena and eliminating one does not necessarily eliminate other. Thus, bloat and over-fitting are important issues while studying generalization ability of evolved GP solutions.

The paper begins by reviewing issues of bloat and over-fitting. The paper classifies different techniques to improve generalization ability of solutions induced by GP into: (i) techniques that minimize evolution of bloated solutions (ii) techniques that minimize evolution of over-fitted solutions.

The next section of the paper discusses issue of bloat, different approaches used by GP practitioners to avoid evolution of bloated solutions and classification of these approaches. Section III discusses issue of over-fitting, different measures of over-fitting and techniques used by GP practitioners to avoid evolution of over-fitted solutions. Section IV presents conclusions.

## II. BLOAT IN GP

The phenomenon of increase in model size without significant gain in terms of fitness is known as bloat. Evolved model with positive bloat means the model is larger than it need to be while negative bloat implies that the model is too small. Bloat has negative effect on performance of GP as large models are computationally expensive to further evolve and are hard to comprehend and have inadequate generalization ability [33].

Application of bloat control schemes give following advantages: (i) produce smaller and interpretable solutions (ii) reduce the search space to regions where good solutions resides (iii) reduce resource consumption by reducing space and time required for evolution and evaluation of solutions (iv) generates more generalize solutions (generalization of solutions tends to decrease as their size increases).While bloat is well defined and can be easily observed there is no common consensus among GP practitioners on why it occurs in GP. Below we discuss five well known theories concerning the reasons why bloat occurs in GP.

Replication Accuracy Theory [15] argues that bloated solutions contain inactive genetic material, called introns. During crossover, swapping can be performed on the inactive genetic material, without affecting effective genetic material of parents. So, fit solutions with more inactive genetic material are less likely to be disrupted by crossover.

Removal Bias Theory [25] observes that inactive genetic material resides in lower portion of GP tree, thus residing in smaller than average size sub-trees. Crossover operation applied to inactive sub-trees generates offspring that has same fitness as its parents. If size of inserted sub-tree is larger than excised sub-tree, the produced offspring retains fitness of parent but gets larger in size than its parent. Thus average solution size of population is increased.

Modification Point Depth Theory [14] extends removal bias theory by observing that there is a correlation between the depth of the node a genetic operator modifies in parent and fitness of produced offspring.

Nature of Program Search Space Theory [13] notes that same solution can be represented by some long as well as some short individuals. Number of ways to represent a solution using long size individuals is high compare to short size individuals. When crossover is unable to produce better solutions, selection gets biased towards solutions that have same fitness as their parents. Since there are more long solutions for a given fitness than short solutions, over a period of time GP drift towards longer solutions.

Crossover Bias Theory [18] explains bloat by assuming that crossover operator on its own does not produce growth or shrinkage in size of solutions. Repeated application of crossover operations push the population towards a particular distribution of tree sizes, where small size trees have high frequency than longer ones. Since small size trees are not useful in solving problem, larger size trees have a selective advantage. Thus, average solution size of population increases.

Different approaches used by GP researchers to overcome the problem of bloat are classified into: (i) Code Editing (ii) Size and Depth Limits (iii) Anti-bloat Genetic Operators (iv) Anti-bloat Selection Schemes. These approaches are presented in Fig. 1. Several bloat control techniques are presented and discussed in [19].

### A. Code Editing / Expression Simplification

GP community used code editing/expression simplification [12] approach to simplify evolved solution by removing redundant code. Code editing can be done before or after evaluation of solution or it can be done at regular interval (generation). However, GP practitioners found that use of this approach can lead to premature convergence [7].

### B. Size and Depth Limit

Koza [12], [33] suggested a method to control growth of models by imposing size and depth limits on generated offspring models. In this method, after application of a genetic operator, the validity test is performed to check if the generated offspring respects the size and depth limit. If the offspring exceeds one of these limits, it is disposed and genetic operation returns best of the selected parents as a result. To estimate the size of a model, poli suggested two steps process: (i) find out minimum possible solution, achievable using given terminal and function sets (ii) add a safety margin of 50%-200% to size of model obtained in previous step  A technique for dynamically adjusting depth and size limits during GP run is proposed in [24]. Authors [24] conclude by experiments that dynamic depth limits produces accurate and smaller models compare to size limits.

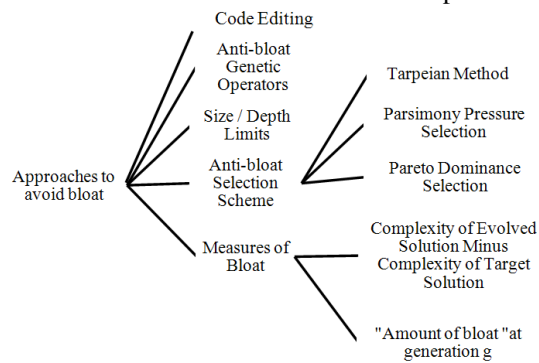

Figure 1 Approaches to Avoid Evolution of Bloated Solutions

### C. Anti-bloat Genetic Operators

Several efforts are made by GP research community in design of genetic operators to control the bloat. For ex. a size fair crossover approach proposed by Crawford et al. [2]. The difference between size-fair crossover and normal crossover lies in selection of second crossover point. The size of sub-tree to be deleted from first parent is used in guiding selection of crossover point in second parent. Thus, the approach applies restriction on the selection of crossover points to prevent the growth.

### D. Anti-bloat Selection Schemes

Tarpeian technique [17] assigns a low fitness value to portion P of individuals having above average size, without evaluating fitness of such individuals. Having low fitness value, selection probability of such individuals for genetic operation will be reduced significantly. An important property of this technique, minimizing the number of evaluations required, differentiates it from other bloat control techniques. Moreover, the technique does not require a priori knowledge of the size of the potential solutions of a problem. However, the technique becomes excessively aggressive in situations where P is large. In this case, the technique rejects a large size individual without considering how fit it is.

A well known approach suggested by Koza [12] to control bloat is parsimony pressure. The approach targets to minimize the rate at which average solution (model) size increases. To achieve this, the approach penalizes fitness (minimizes the selection probability) of solution based on its size. Each solution k is assigned a new selection fitness $f_{sel}(k) = f(k) - c*s(k)$, where f(k) is the original fitness, s(k) is the size of solution k and c is the parsimony coefficient. Selection of right value of parsimony coefficient is very important as it decides the intensity with which the bloat is controlled. If the coefficient is set to very low value then there is no force to minimize the bloat. On the other end, if it is set to large value then runs will evolve extremely small but inaccurate solutions by neglecting the main goal of optimization of fitness. Choosing right value of parsimony coefficient is difficult and depends on problem to be solved. Using constant value for parsimony coefficient can only achieve partial control over average size of solution over a period of time [20]. Co-variant parsimony pressure approach sets value of parsimony coefficient dynamically during evolutionary run, is proposed in [20]. [20] concludes that the method achieves tight control over average size of solutions.

Parsimony pressure selection combines two objectives, size and fitness, into a single objective, whereas multi-objective selection keeps the two objectives separate. Multi-objective selection approaches uses concept of pareto-dominance optimization scheme. In pareto-dominance optimization, an individual X is said to dominate individual Y if X is as good as Y in all objectives and is better than Y in at least one objective. Pareto-dominance selection scheme generates a set of acceptable trade-off optimal solutions. This set is referred as a Pareto set. A modified tournament selection operator based on pareto dominance is proposed in [4]. The operator selects a solution only if it is not dominated by a set of randomly chosen solutions.

TABLE I CLASSIFICATION OF BLOAT CONTROL APPROACHES

| Characteristics | Bloat Control Approaches | | | | | | | |
|---|---|---|---|---|---|---|---|---|
| | Code Editing | Size / Depth Limit | Anti-Bloat Gen. Op. | Tarpeian | Linear Parsimony | Co-variant Parsimony | Multi-objective selection | Operator Equalization |
| Direct / Indirect | Direct | Indirect | Indirect | Indirect | Indirect | Indirect | Indirect | Indirect |
| Parametric / Non-Parametric | Non-parametric | Parametric | Parametric | Parametric | Parametric | Parametric | Non-parametric | Parametric |
| Adaptive / Non-Adaptive | Non-Adaptive | Non-Adaptive | Non-Adaptive | Non-Adaptive | Non-Adaptive | Adaptive | Non-Adaptive | Non-Adaptive |
| Phase of GP | Other | Breeding | Breeding | Evaluation | Evaluation | Evaluation | Selection | Breeding |

*E. Operator Equalisation*

Operator equalisation technique [3] controls bloat by biasing search towards smaller or larger individuals. User has to specify solution length distribution that she wish GP system should use while sampling solution space. The technique controls sampling rates of specific solution lengths by probabilistically accepting each newly produced individual (solution) based on its length.

*F. Bloat Measures*

Amount of bloat is measured based on relationship between average model length growth and average fitness improvement at current generation compared to respective values at generation zero in [30]. The measure hypothesize that there is no bloat at generation zero. Amount of bloat is computed by taking the difference of structural complexities of evolved solution and target solution in. [22]

We classify different bloat control approaches into: (i) Direct/Indirect [28]: Direct approaches control bloat by simplifying the solutions using special operators. Code editing approach is an example of direct bloat control approach. Indirect approaches control bloat by accepting or rejecting the solutions modified by genetic operators or through selection. (ii) Parametric/Non-Parametric [23]: Parametric parsimony pressure schemes evaluate final fitness of an individual using a parametric model comprising of raw fitness and size of an individual. Size/Depth limits and tarpeian approaches are example of parametric bloat control approach. (iii) Adaptive/Non-Adaptive [28]: Depending on whether the intensity of parsimony pressure (value of parsimony coefficient) is fixed or vary during the GP run, bloat control approaches are classified into adaptive or non-adaptive. Covariant parsimony pressure is an example of non-adaptive bloat control approach. (iv) Phase of GP [23]: Depending on the phase of GP at which bloat control method applies, bloat control approaches can be classified. Size/depth limit and anti-bloat genetic operators bloat control approaches are applied at the breeding phase of GP. Table I presents classification of bloat control approaches.

III. OVER-FITTING IN GP

Development of an unknown model from finite training data carries a risk of excessively fitting model to data. This phenomenon is known as over-fitting in the field of data based modeling. The over-fitted model tries to model noise present in training data rather than explaining the whole training data. The over-fitted models have properties of low training errors and high generalization errors.

Different approaches used by GP practitioners to avoid evolution of over-fitted models are: (i) Interval Arithmetic (ii) Partitioning Data (iii) Reducing Complexity of Models (iv) Ensemble of Heterogeneous Models (v) Multi-objective Optimization (vi) "Linear Scaling" with "No Same Mate" selection. These approaches are presented in Fig. 2.

### A. Interval Arithmetic

It is important to make sure that evolved models do not have an undefined (asymptotic, infinity) behaviour in their output for unseen input data points. Usually protected operators are used in GP to avoid an undefined behaviour of evolved model at unseen input data points. For ex. Division by zero or taking square root of a negative number may produce an undefined behaviour. Three different approaches to avoid this situation: (i) use of ad-hoc values to avoid the undefined behaviour, proposed by Koza [12] (ii) removal of evolved model that has an undefined behaviour from population (iii) restricting function set to contain only those functions that do not produce any undefined behaviour.

Use of first two approaches ensures evolution of well-behaved models on the training dataset, but it is still possible that the evolved models may have undefined behaviour on data-points that are not covered by the training dataset. Use of interval arithmetic is proposed in [10], [11] to evolve reliable models that do not have undefined behaviour in their output range. The method calculates output bound recursively for every node of the model, given the bounds of the input arguments. The models comprising nodes having undefined values for output bound are identified and can be removed from the population. Interval arithmetic is used in [27], to ensure robustness of evolved models using symbolic regression through simulated annealing.

### B. Partitioning Data

Hold-out Method divides the available data into two disjoint data sets – training data set and test data set [6], [33]. Training data set is used to evolve the model and the test data set is used to approximate the generalization ability of the evolved model. The over-fitted models can be easily identified by the fact that they reveal very good fitness on training data set but poor fitness on test data set.

N-Fold Cross Validation method divides available data into N disjoint parts. Model training will be done N times, each time using N−1 parts as training data and remaining part as test data [33]. N-fold cross validation is not suited for empirical modeling using GP because for each of N training run, the algorithm can induce a different model.

To use GP for data based modeling, a preferable approach is to divide the available data into three parts – training data set, validation data set and test data set. Training data set is used to evaluate the fitness of the models, where as the validation data set is used to find out the over-fitted models from the evolved models. The validation data set is useful in selecting models, where as the test data set is used to estimate the generalization error of selected models on unseen data. Validation data set is used in [6], [22], [33] to distinguish between over-fit solutions and exact solutions.

A measure that computes over-fitting of a model by obtaining relationship between model's fitness on the training set and test set is proposed in [30]. The proposed idea is based on following rules: (i) if model's fitness on test set is better than it's fitness on training set then there is no over-fitting. (ii) if model's fitness on test set is better than fitness of best model, found so far, on test set then there is no over-fitting. (iii) otherwise, amount of over-fitting is computed by taking difference of model's fitnesses on training set and test set at current generation and difference of training and test fitnesses of best models found so far. The drawback of this measure is that it depends on how training and test data set are selected.

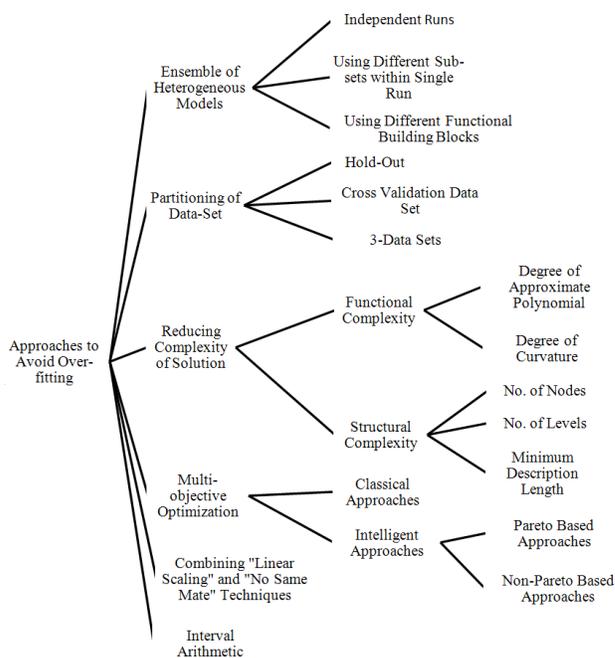

Figure 2 Approaches to Avoid Evolution of Over-Fitted Solutions

### C. Reducing Complexity of Models

A large size (complex) model than required is of little practical use and hard to interpret. This fact is reflected in Occam's Razor principle that tells that between models of comparable quality, simpler model is preferred over complex one. To reduce over-fitting and to improve interpretability of models, evolutionary process must control complexity of models and favour simpler models during evolution.

Complexity of an evolved GP model can be measured in genotype space or in phenotype space. In general, measuring complexity in one space is equivalent to measuring it in other space. However, for some problem classes this is not true and for such problems measuring phenotypic complexity is preferred in research community. Different kinds of complexities associated with every model are: (i) structural complexity of model which emphasis on compactness of genotype (ii) behavioral complexity of model which emphasis on smoothness of phenotype [31].

*1) Structural Complexity*

Different measures used to measure structural complexity of solution are: (i) Number of nodes in a tree (ii) Number of levels in a tree (iii) Minimum description length (iv) Expressional complexity of a model, determined by sum of number of nodes in all sub-trees of a given model.

*2) Behavioral Complexity*

Many GP researchers believe that issue of over-fitting is linked with the functional complexity of the solution. Functional complexity of a model is measured by computing model's behaviour (output) over possible input space. A new complexity measure, called, order of nonlinearity of a model, to favour smooth behaviour of response surface and to deject highly nonlinear (unstable) behaviour is proposed in [31]. The order of nonlinearity of a model is measured by approximating minimal degree of polynomial necessary to approximate the model. The concept behind the proposed measure is that over-fitted models are approximated by polynomial of high degree due to high oscillation in their behaviour [27]. Parsimony pressure approach is suggested in [6] to reduce the complexity of models and thus to improve the generalization ability of models.

A complexity measure based on slope of line segments is proposed in [29]. The slope-based functional complexity (SFC) is computed by taking sum of differences of slope of consecutive line segments. However, authors [29] calculated SFC measure for each problem dimensions separately in case of multi-dimensional problems. To overcome limitations of SFC, a new measure based on concept of measuring amount of variation in output is presented in [29].

### D. Ensemble of Heterogeneous Models

Averaging the output of diverse models to improve prediction accuracy and using their consensus to assess the trust is proposed in [11]. Different strategies to generate robust and diverse models [11] are: (i) using different function sets (ii) executing independent runs (iii) using different subsets for each generation within a single evolution. Advantage of these strategies is that the whole available dataset is used for model development compared to traditional approach of dividing the dataset into training, test and validation subsets to mitigate risk of over-fitting.

### E. Multi-Objective Optimization

Real world problems frequently demands to satisfy multiple and conflicting objectives. For ex. finding vehicle that can travel maximum distance in a day while consuming least energy is a multi-objective optimization problem [16]. The aim of multi-objective optimization is to produce set of acceptable trade-off optimal solutions. Two different approaches to solve multi-objective optimization problems are: (i) classical approaches (ii) intelligent approaches.

*1) Classical Approaches*

Solving multi-objective optimization problems using classical approaches convert multiple objectives into a single objective. The conversion of multiple objectives into single objective is done either by aggregating all objectives in a weighted function or optimizing one objective and considering others as constraints. The approach has following limitations: (i) requires a priori preferential information about objectives (ii) the aggregated function produces a single solution (iii) trade-offs between objectives cannot be assessed easily [16]. One approach, weighted aggregation, converts multi-objective optimization problem into a single objective optimization problem by applying a weighted function to objective vector. It requires a priori knowledge of relative importance of different objectives. In absence of such knowledge, selection of weights can be problematic. Dynamic Weighted Aggregation [9] solves the problem by changing the weights incrementally.

*2) Intelligent Approaches*

These approaches seek for simultaneous optimization of individual objectives compared to single objective optimization of aggregation-based techniques. Intelligent approaches are classified into: (i) Non-pareto based approaches (ii) Pareto based approaches. Difference between two lies in the fact that later approach use pareto-ranking of models to find out the probability of replication of a model.

Vector Evaluated Genetic Algorithm (VEGA) [21] is a Non-pareto based approach. The selection scheme of VEGA partitions whole population into as many equal size sub-parts as there are objectives. Then selection of fittest individuals for each objective from these sub-parts is performed. The drawback of VEGA is that it generates models those are optimal in one of the objectives and not truly pareto optimal.

Pareto based approaches use sorting of non-dominated solutions along with a niching mechanism to avoid premature convergence. The fitness of a solution is determined by its dominance in the population. The niching mechanism is used to maintain diversity among solutions. The way fitness value of a solution is calculated is used to differentiate between these approaches. These techniques are: (a) Multi Objective Genetic Algorithm [5] calculates fitness of a solution based on the number of other solutions it dominates. (b) Non-dominated Sorting Genetic Algorithm [26] classifies population on basis of non-dominance before applying selection step. (c) Niched Pareto Genetic Algorithm [8] applies tournament selection based on pareto dominance. (d) Strength Pareto Evolutionary Algorithm [32] uses an external archive to maintain non-dominated solutions found in previous generations. Fitness of a solution depends on the solutions stored in the external archive.

### F. "Linear Scaling" with "No Same Mate" Selection

Costelloe and Ryan [1] experimentally concluded that GP with linear scaling may perform better compared to standard GP on training data, but the technique does not generalize well on test data. They proposed to combine "No Same

Mate" selection with linear scaling to improve generalization ability of evolved GP solutions.

IV. CONCLUSIONS

This paper discussed problem of generalization ability of GP solutions. The paper presents two issues: bloat and overfitting related to generalization ability of GP solutions. The paper summarized state of the art approaches used by GP practitioners to control evolution of bloated solutions and classifies them into: Direct/Indirect, Parametric/Non-Parametric, Adaptive/Non-Adaptive. The paper also reviewed different approaches to reduce evolution of overfitted models. The paper presents advantage and disadvantage of different generalization approaches with which GP practitioners must be aware of. This will help them in selection of better generalization approach suited for solving specific empirical modeling problem using GP.